\documentclass[11pt]{article}

\usepackage[final]{acl}

\usepackage{times}
\usepackage{latexsym}

\usepackage[T1]{fontenc}

\usepackage[utf8]{inputenc}

\usepackage{microtype}

\usepackage{inconsolata}

\usepackage{graphicx}

\usepackage{amsmath,amssymb,amsfonts}
\usepackage{booktabs}
\usepackage{enumitem}
\usepackage{xcolor}
\usepackage{caption}
\usepackage{subcaption}
\usepackage{wrapfig}
\usepackage{mathtools}
\usepackage[most]{tcolorbox}
\usepackage{forest}
\usepackage{adjustbox}
\usepackage{tikz}
\usetikzlibrary{trees,positioning,shapes,shadows,arrows.meta}

\definecolor{cat}{HTML}{0B6FA4}
\definecolor{sub}{HTML}{3FA2DA}

\forestset{
tax/.style={
for tree={
draw, rounded corners, align=center, inner sep=3pt, outer sep=2pt,
edge={-latex},
s sep=7pt, l sep=8pt,
font=\small
}
},
cat/.style={fill=cat!15, draw=cat},
subcat/.style={fill=sub!15, draw=sub}
}

\title{Barriers to Discrete Reasoning with Transformers: A Survey Across Depth, Exactness, and Bandwidth}

\author{
Michelle  Yuan, Weiyi Sun, Amir H. Rezaeian, Jyotika Singh,\\
\textbf{Sandip Ghoshal,} \textbf{Yao-Ting Wang,} \textbf{Miguel Ballesteros,} \textbf{Yassine Benajiba}\\
Oracle AI\\
\textbf{Correspondence:} \href{mailto:email@domain}{michelle.yuan@oracle.com}
}

\begin{document}
\maketitle

\begin{abstract}

Transformers have become the foundational architecture for a broad spectrum of sequence modeling applications, underpinning state-of-the-art systems in natural language processing, vision, and beyond. However, their theoretical limitations in discrete reasoning tasks, such as arithmetic, logical inference, and algorithmic composition, remain a critical open problem. In this survey, we synthesize recent studies from three theoretical perspectives: circuit complexity, approximation theory, and communication complexity, to clarify the structural and computational barriers that transformers face when performing symbolic computations. By connecting these established theoretical frameworks, we provide an accessible and unified account of why current transformer architectures struggle to implement exact discrete algorithms, even as they excel at pattern matching and interpolation. We review key definitions, seminal results, and illustrative examples, highlighting challenges such as depth constraints, difficulty approximating discontinuities, and bottlenecks in inter-token communication. Finally, we discuss implications for model design and suggest promising directions for overcoming these foundational limitations.

\end{abstract}

\section{Introduction}

Transformer architectures~\cite{vaswani2017attention} have driven remarkable progress across language understanding~\cite{raffel2020exploring}, generation~\cite{achiam2023gpt,guo2025deepseek}, and perception tasks~\cite{dosovitskiy2020image,kirillov2023segment}, setting new performance benchmarks and enabling a wide range of applications~\cite{tunstall2022natural,Snh2023}. These large models have been coined as ``foundation models'' because of their consequential impact for society~\cite{bommasani2022opportunitiesrisksfoundationmodels}. Despite these advances, a persistent gap remains in their ability to perform reliable discrete reasoning, such as precisely solving arithmetic expressions~\cite{lee2023teaching}, evaluating logical formulas~\cite{dziri2023faith}, or generalizing systematic rules to longer or unfamiliar examples~\cite{wu2023counterfactual,gao2024insights,yang2024efficient, 11218142, allamanis2025disprovingprogramequivalencellms, singh-etal-2025-llms}.

This survey examines the fundamental factors that constrain transformer-based models in discrete reasoning domains. Rather than focusing on empirical evaluation or incremental improvements, we systematically review and synthesize results from three major theoretical disciplines:

\begin{enumerate}[itemsep=0pt, parsep=0pt]
    \item \textbf{Circuit Complexity} frames model capacity in terms of computational depth and parallelism, emphasizing why certain algorithmic processes fundamentally require sequential computation
    \item \textbf{Approximation Theory} exposes the inherent difficulties in representing discontinuous, rule-based functions as transformers are trained for continuous mappings
    \item \textbf{Communication Complexity} formalizes the limits of information exchange within self-attention and elucidates how architectural choices restrict multi-hop reasoning and coordination across tokens.
\end{enumerate}

Through this synthesis, we provide readers with a cohesive understanding of why transformers succeed in interpolation tasks (e.g. summarization) but fall short in reliably executing symbolic algorithms. Through outlining key definitions, representative tasks, and landmark theoretical results, we contextualize both the current abilities and intrinsic limitations of transformer architectures. We also identify promising avenues for future research, including hybrid models that combine neural and symbolic components or extend transformer architectures beyond their current design points.

\begin{figure*}
    \centering
    
\tikzset{
    basic/.style  = {draw, text width=3cm, align=center, font=\sffamily, rectangle},
    root/.style   = {basic, rounded corners=2pt, thin, align=center, fill=green!30},
    onode/.style = {basic, thin, rounded corners=2pt, align=center, fill=green!60, text width=3cm,},
    tnode/.style = {basic, thin, align=left, fill=pink!60, text width=15em, align=center},
    xnode/.style = {basic, thin, rounded corners=2pt, align=center, fill=blue!20, text width=5cm,},
    wnode/.style = {basic, thin, align=left, fill=pink!10!blue!80!red!10, text width=6.5em},
    edge from parent/.style={draw=black, edge from parent fork right}
}

\begin{forest} 
for tree={
    grow=east,
    growth parent anchor=west,
    parent anchor=east,
    child anchor=west,
    edge={->, >={latex}},
    l sep=10mm,
}
[Barriers to Discrete Reasoning with Transformers, basic
    [Communication Complexity, xnode
        [Sequential Augmentation, tnode]
        [Long-range Comparison, tnode]
        [Sequential Relay, tnode]
    ]
    [Approximation Theory, xnode
        [Unbounded Worst-case Errors, tnode]
        [Approximating Multi-step Problems, tnode]
        [Approximating Piecewise-Constant Functions, tnode]
    ]
    [Circuit Complexity, xnode
        [Constant Output Length, tnode]
        [Constant Depth, tnode]
        [Finite Numerical Precision, tnode]
    ]
]
\end{forest}
    \caption{Taxonomy of barriers to discrete reasoning in transformers, organized by theoretical lens: circuit complexity, approximation theory, and communication complexity.}
    \label{fig:taxonomy}
\end{figure*}
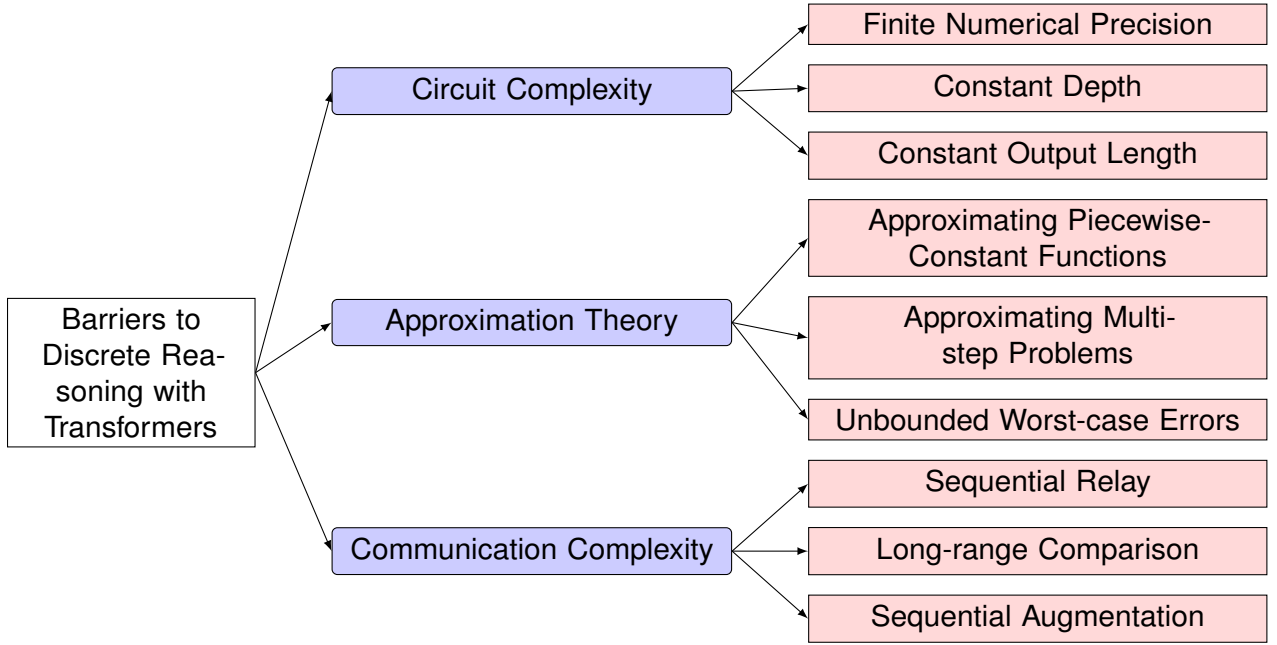

This survey is structured as follows: we first provide necessary background on transformers and discrete reasoning tasks, then dedicate individual sections to each of the three theoretical frameworks. Figure~\ref{fig:taxonomy} shows an outline of the issues and subtopics for each framework. We conclude by drawing connections between these perspectives, surveying related work, and outlining open problems and opportunities for the next generation of reasoning-capable machine learning systems.

\section{Background}

This section provides essential context on the architectural design of transformers and clarifies why discrete reasoning tasks pose unique challenges for modern neural models. We begin by outlining the key components and mechanisms of transformer architectures, followed by an overview of the types of discrete reasoning problems that motivate our analysis. Finally, we introduce three foundational theoretical frameworks, circuit complexity, approximation theory, and communication complexity, that will anchor the discussion in subsequent sections.

\subsection{Transformer Architectures}

Transformers~\cite{vaswani2017attention} have revolutionized sequence modeling by replacing recurrent layers with a purely attention-based design. The fundamental architecture comprises multiple stacked layers, each containing a multi-head self-attention mechanism and a position-wise feedforward sublayer, typically augmented with residual connections and layer normalization. This parallelized design better captures long-range dependencies compared to previous NLP architectures~\cite{devlin2019bert}.
Later on, researchers have looked towards scaling model size~\cite{hoffmann2022training}. This involved increasing both the depth (the number of stacked layers) and the width (hidden size and attention heads), leading directly to the development of modern LLMs~\cite{achiam2023gpt}. However, this scaling strategy quickly exposed two critical limitations~\cite{duman2023computational}. First, self-attention scales quadratically with the input sequence length. Second, at inference time, the fixed number of stacked layers dictates a fixed sequential depth, which imposes theoretical bounds on the number of computational steps the model can perform in a single forward pass.

To overcome these limitations, LLMs have adopted various architectural and algorithmic innovations. To efficiently scale parameter count while managing inference cost, the Mixture-of-Experts layer~\cite{zhou2022mixture} allows the creation of models with trillions of parameters while only activating a small fraction for any given input. Memory-efficient techniques like Sparse Attention~\cite{tay2020sparse} and Flash Attention~\cite{dao2022flashattention} are developed to address the quadratic complexity of long inputs. Furthermore, modifications to positional encoding, such as RoPE, where positional information is encoded by rotating the query and key vectors, enhance the models' ability to maintain and model dependencies effectively over increasingly longer sequences~\cite{su2024roformer}.
While these innovations improve efficiency, they do not inherently increase the fixed computation depth (Section~\ref{sec:circuit}), guarantee symbolic precision (Section~\ref{sec:approx}), nor fully address memory bandwidth (Section~\ref{sec:communication_complexity}).

\subsection{Discrete Reasoning Tasks}

Discrete reasoning involves operations that require precise, rule-based manipulation of symbolic entities, often with exact correctness required for each computational step. Examples include arithmetic problems, Boolean logic evaluation, and parsing. Unlike tasks involving perceptual or semantic similarity, discrete reasoning is characterized by sharp decision boundaries, combinatorial complexity, and non-smooth target mappings. Transformers often rely on similarity-based heuristics learned from data, rather than internalizing algorithmic procedures or conditional logic~\cite{mirzadeh2024gsmsym}. This distinction lies at the heart of their observed failure modes on problems that require compositionality, systematic generalization, or step-wise execution of rules~\cite{dziri2023faith}.

Recent studies continue to report notable shortcomings of foundation models on discrete reasoning tasks. The MathArena Apex leaderboard~\cite{balunovic2025matharena} contains challenging combinatorics problems like: 
\begin{quote}
\textbf{The Zigzagging Chessboard.}
Let $P$ be a polygon formed by the edges of an infinite chessboard, which does not intersect itself. Let $a_1$, $a_2$, and $a_3$ denote the number of unit squares that have exactly 1, 2, or 3 edges on the boundary of $P$, respectively. Find the largest real number $k$ such that the inequality $a_1 + a_2 > k a_3$ holds for each polygon constructed with these conditions.
\end{quote}
This problem requires discrete reasoning because it involves deducing relationships based on the exact combinatorial arrangement and counting of unit squares with specific properties on an infinite chessboard. On the MathArena Apex leaderboard, Gemini 3 Pro currently has the highest accuracy of 23.44\%. On other related benchmarks like FrontierMath Tier-4~\cite{glazer2024frontiermath}, GPT 5.2 Pro has the highest accuracy of 29.2\%. The low performance from these frontier models reflects the intrinsic difficulty posed by combinatorial structures and discrete reasoning in the benchmarks.

\subsection{Theoretical Frameworks}

\paragraph{Circuit Complexity}
Circuit complexity is a branch of computational complexity theory concerned with the resources required to compute functions using Boolean circuits. A Boolean circuit can be formally described as a directed acyclic graph in which each internal node represents a logic gate (such as AND, OR, NOT), and the leaves correspond to input variables $x_1, x_2, \ldots, x_n$. The primary measures of a circuit are its \emph{size} (the total number of gates, denoted $S(f)$ for a function $f$) and its \emph{depth} (the length of the longest path from an input node to the output, denoted $D(f)$). Circuit classes such as $\mathrm{AC}^0$, $\mathrm{TC}^0$, and $\mathrm{NC}^1$ categorize families of functions based on allowable depth, gate types, and fan-in. 

The class AC$^0$ consists of all Boolean circuits with constant depth, unbounded fan-in for AND and NOT gates, and polynomial size. Building on this, the class TC$^0$ extends AC$^0$ by including majority gates, which are capable of processing unbounded inputs and outputting 1 if and only if the majority of those inputs are 1. Another important class, NC$^1$, is similar to AC$^0$ but restricts circuits to logarithmic depth (specifically O(logn)) and bounded fan-in. The known relationships are $\text{AC}^0 \subset \text{TC}^0 \subseteq \text{NC}^1$ in DLOGTIME-uniform, L-uniform, and non-uniform settings.
Key questions in circuit complexity include establishing lower and upper bounds for the resources needed to compute specific Boolean functions~\cite{heribert1999circuit}.

\paragraph{Approximation Theory}
Approximation theory studies how well classes of functions (such as polynomials or neural networks) can approximate target functions, usually with respect to some norm or distance metric. Given a target function $f: X \to Y$ and a family of approximating functions $\mathcal{F}$, the goal is to find an $f^* \in \mathcal{F}$ that minimizes the approximation error $\|f - f^*\|$ according to a specified norm, such as the $L^\infty$ (uniform) or $L^2$ norm. Classical results, such as the Universal Approximation Theorem~\cite{hornik1989multilayer}, assert that for certain families of functions (e.g., sufficiently wide neural networks with non-polynomial activation functions), it is possible to approximate all continuous functions on compact domains arbitrarily well. However, approximation theory also characterizes the inherent difficulty of approximating discontinuous functions, where the error depends on factors like the smoothness and complexity of the target, and the expressive power or depth of the approximating family~\cite{devore1998nonlinear}.

\paragraph{Communication Complexity}
Communication complexity analyzes the amount of communication required between two or more parties to collaboratively compute a function whose input is distributed among them. Instead of focusing on computational resources, this framework quantifies the minimal number of bits that must be exchanged for the correct computation of a function $f(x, y)$, where $x$ is known to Alice and $y$ to Bob, and both aim to compute $f(x, y)$ with the least amount of communication. The deterministic communication complexity $D(f)$ of a function $f$ is defined as the minimum number of bits exchanged between the parties in the worst case, using an optimal protocol. Extensions consider randomized protocols, multiple rounds of communication, or multiparty variants. Canonical problems studied in communication complexity include the Equality function, Disjointness, and Greater-Than, each exhibiting different lower and upper bounds for bit and round complexity~\cite{kushilevitz1997communication}.

\section{Circuit Complexity: Why Depth Becomes a Bottleneck}
\label{sec:circuit}
Recent theoretical work has sought to understand Transformer's expressivity using tools from circuit complexity and formal languages. This line of work provides a principled account of which discrete reasoning tasks transformers can or cannot perform, and clarifies how different transformer variants change the network's representation power. In this section, we focus on how finite precision, constant depth, and constant output constraints each create distinct barriers to discrete reasoning, but we also highlight how relaxing these assumptions can expand transformer capability. We build on unifying frameworks such as \citet{strobl2024formal}.

\subsection{Finite Numerical Precision}
In practice, transformer models operate under finite-precision arithmetic, mapping to threshold circuits with limited numerical granularity~\cite{merrill2023logic,merrill2023parallelism,merrill2022saturated}. These models are fundamentally limited in their ability to represent functions requiring precise bitwise logic, such as parity or majority. For example, encoder-only transformers with hard attention correspond to $\mathsf{AC}^{0}$ circuits and are unable to recognize parity, majority, or DYCK-$k$ languages~\cite{hao2022formal}. While softening attention mechanisms (e.g. average-hard or softmax) extends expressivity slightly beyond $\mathsf{AC}^{0}$~\cite{merrill2022saturated,bhattamishra2020ability,yao2021self,chiang2022overcoming}, and the implementation of embedding schemes such as RoPE do not surpass the upper bound of DLOGTIME-uniform $\mathsf{TC}^{0}$~\cite{merrill2023parallelism,chen2024rope}. However, if infinite precision were possible, one could in theory go well beyond these bounds, though this is unrealistic in current architectures.

\subsection{Constant Depth}
Transformers are commonly deployed with a constant number of layers, reflecting a fixed computational depth that sharply limits their ability to perform deeply compositional or sequential tasks. Classic and recent results show that finite-depth, finite-precision transformers without intermediate decoding have small-circuit upper bounds, confining their expressivity to uniform $\mathsf{TC}^{0}$ or below~\cite{merrill2023parallelism,merrill2023logic,penglimitations,merrill2022saturated,zubiclimits}. They cannot, for instance, compute multi-digit addition, recognize nested formal languages, or resolve long-range dependencies. However, if one allows the number of layers to scale with the input, or augments the model with additional programmatic mechanisms (such as auxiliary memory or scratchpads), then in principle these barriers can be overcome~\cite{perez2021attention,qiuask}. 

\subsection{Constant Output Length (No Sequential Augmentation)}
A further constraint arises when transformers are limited to producing outputs of constant length, precluding stepwise, intermediate reasoning. Without chain-of-thought (CoT) or scratchpad mechanisms, all computation must occur in a single forward pass, restricting models to shallow circuits and causing systematic underperformance on multi-step or compositional tasks~\cite{feng2023towards}. Recent theory shows this is an inherent barrier, but if transformers are permitted to emit longer outputs, effectively augmenting depth with CoT tokens, their expressivity hierarchy expands dramatically. \citet{merrill2023expressive} show that logarithmic-length CoT yields modest gains, linear chain-of-thought suffices for all regular languages, and with polynomial CoT or certain architectural variants, transformers \citet{li2024chain} demonstrate that with $T$ CoT steps, a constant-depth, constant-precision transformer can simulate any Boolean circuit of size $T$. \citet{qiuask} further prove that, by providing suitable program-prompts, a fixed-size decoder-only transformer can attain Turing-complete expressive capacity by emitting $O(t(n))$ chain-of-thought, making output length a true computational resource. 

\begin{tcolorbox}[colback=yellow!15!white, colframe=orange!80!black, title=Summary]
Circuit complexity theory clarifies that transformers are provably unable to represent many essential symbolic functions under realistic settings, such as finite precision, constant architectural depth, and no access to chain-of-thought. Nonetheless, relaxing any of these constraints, by increasing precision, depth, or enabling intermediate outputs, yields substantial gains in representational power. This suggests concrete directions for future architecture and prompt design to enhance discrete reasoning.
\end{tcolorbox}
\section{Approximation Theory: Why Exactness is Hard}
\label{sec:approx}

The Universal Approximation Theorem (UAT) states that sufficiently large neural networks can approximate any continuous function on a compact domain~\cite{hornik1989multilayer}. However, this property does not extend to discontinuous functions, which are the very functions that dominate discrete reasoning tasks. Symbolic tasks frequently have outputs that change abruptly at precise input thresholds. Approximating such sharp boundaries with smooth neural networks introduces transition regions, resulting either in systematic errors (if smoothed) or excessive sensitivity (if made steep)~\cite{arora2022understanding,chen2023does}.
Discrete reasoning tasks naturally extend over unbounded spaces, i.e., $f: \mathbb{N}^n \to \mathbb{N}$ for $n$-digit addition, violating the UAT's compactness assumption. Thus, transformers trained as smooth interpolators cannot guarantee uniform error control as sequence or input length grows~\cite{hahn2024sensitive}.

Note that the limitations mentioned in this section are not unique to transformer architectures. In fact, the constraints described here are representative of broader classes of inductively-trained neural networks, whose ability to learn target functions is fundamentally limited by principles established in classical learning theory. For instance, learnability theory formally characterizes which function classes can be efficiently learned and highlights the inherent difficulty of precisely modeling discontinuous or highly complex rules from finite data~\cite{valiant1984theory}.

\subsection{Difficulty Approximating Piecewise-Constant Functions}

Symbolic functions are often \emph{piecewise-constant}. Take $n$-digit addition as an example. The function mapping $(a, b) \mapsto s$ (where $s$ is the sum sequence) is constant on most of the input space but jumps sharply at digit-wise carry thresholds. Formally, let $c_i = 1$ if $a_i + b_i + c_{i-1} \geq 10$ (carry at position $i$), where $a = (a_1, \dots, a_n)$ and $b = (b_1, \dots, b_n)$. The set of discontinuity surfaces is
\[
\mathcal{D} = \{(a, b) \mid a_i + b_i + c_{i-1} = 10 \text{ for some } i \},
\]
which grows linearly with $n$. For vectors near $\mathcal{D}$, small perturbations in a single digit flip the carry and, recursively, all subsequent digits.

Empirical analyses show that transformer display sharp accuracy drops on discontinuities. For example, \citet{hu2024case} observe regression in carry-over operations on multi-digit addition and multiplication, and conclude that transformers rely on case-based reasoning to match unseen examples to previously seen problems. Visualizing the model failures will represent these discontinuities as ``holes''. \citet{zhao2024exploring} specifically design discontinuities as ``traps'' in math word problems and demonstrate that state-of-the-art LLMs cannot solve these augmented problems.

\subsection{Compounding Errors for Multi-step Problems}

The complexity intensifies with the number and composition of discontinuities. Let $f^{(k)}$ denote $k$-fold function composition (e.g., $k$ sequential carries in addition or $k$ nested logical operations). For each composition step, the Lipschitz constant or associated parameter norm must grow to retain precision at boundaries, i.e.,
\[
\left\| f^{(k)}(x) - \hat{f}^{(k)}(x) \right\| \leq L^k \varepsilon 
\]
for some $L > 1$, so error compounds exponentially in the number of steps unless the per-step error is super-polynomially small. In practice, fixed-depth neural networks cannot achieve this for increasing $k$~\cite{bubeck2021universal}.

\citet{lee2023teaching} run extensive experiments on length generalization for multi-digit arithmetic problems. Despite various training ablations, which includes data sampling, data formatting, and optimization strategies, transformers are unable to generalize to problems with more computational steps. \cite{dziri2023faith} also observe the same regression as they increase the number of steps in multi-step reasoning problems. 


\subsection{Unbounded Worst-case Error for Discontinuities}
\label{ssec:approx-catastrophic}

Training strategies such as label smoothing~\cite{muller2019does} or calibration~\cite{desai2020calibration,chen2023calibrating} can improve mean squared error locally but do not address the exponential number of discontinuity boundaries for deeper compositions. Let $E_{\mathrm{avg}}$ and $E_{\mathrm{max}}$ denote average and worst-case error, respectively:
\[
E_{\mathrm{avg}} = \mathbb{E}_{x \sim P_{\text{train}}} \left[\, |f(x) - \hat{f}(x)|\, \right]
\]
\[
E_{\mathrm{max}} = \sup_{x \in \mathcal{X}} |f(x) - \hat{f}(x)|
\]
While $E_{\mathrm{avg}}$ may be small, $E_{\mathrm{max}}$ remains large, especially for out-of-distribution examples. \citet{wu2023counterfactual} 
 introduce a framework to evaluate ``counterfactual'' reasoning where the logical rules are transferred to a counterfactual world and show that large transformers fail to generalize to alternative domains. There is a whole line of work that shows transformers susceptible to simple, adversarial attacks~\cite{guo2021gradient, shayegani2023survey, gao2024attacking}. In the end, transformers will show a ``bimodal'' error -- many exact outputs, with rare but catastrophic failures on decision boundaries.


\begin{tcolorbox}[colback=yellow!15!white, colframe=orange!80!black, title=Summary]
Approximation theory shows that the intrinsic smoothness bias of transformers conflicts with the highly discontinuous, rule-based structure of algorithmic reasoning. Increasing model width or dataset size cannot overcome the compounding effect of boundaries and error growth in deep compositions. New architectures or training methods targeting discontinuity and composition may be necessary for robust symbolic reasoning.
\end{tcolorbox}

\section{Communication Complexity: Why Bandwidth and Rounds Matter}
\label{sec:communication_complexity}

Communication complexity offers a distinct perspective on the foundational limitations of transformers in discrete reasoning. While previous sections explored the role of depth and approximation theory, here we focus on how the limits of message-passing and bandwidth within the self-attention architecture sharply constrain multi-step and symbolic computation.

\subsection{Sequential Relay Limits}
Transformers implement a highly parallel message-passing scheme: in every self-attention layer (``round''), each token can broadcast information to every other token by updating its representation based on a mixture of all sequence elements. The model’s depth directly determines the number of these broadcast rounds it can perform, while the hidden dimension of each token controls the ``bandwidth'' of each individual message. This structural design is efficient for pattern recognition over moderately-sized contexts but creates bottlenecks for problems requiring several sequential steps of reasoning or information relay along chains of tokens. Practically, increasing hidden size improves the fidelity of exchanged messages but cannot substitute for more architectural rounds. For many sequential tasks there are provable depth-width trade-offs, namely constant (or small) depth transformers must increase model dimension polynomially in input length to perform $k$-step sequential composition exactly~\cite{chen2024multilayer, keles2022selfattn}. 

\paragraph{Example: Long-carry Propagation as Pointer Chasing.} 
Consider the task of summing two $n$-bit binary numbers, $a$ and $b$, where the output most significant bit (MSB) depends on the correct propagation of carries from the least significant bit (LSB) through all intermediate positions. Each digit position can be mapped to a ``party'' holding its local inputs $(a_i, b_i)$, with the value of the carry bit at position $i+1$, $c_{i+1}$, depending on both the carry $c_i$ and the local inputs~\cite{10.5555/3666122.3667718, sanford2024transformers, peng2024on}. To compute the MSB exactly, information about the final carry must propagate sequentially from the LSB, traversing all $n-1$ positions, a classic ``pointer chasing'' problem~\cite{10.1145/301250.301413}. Within a transformer, each layer corresponds to just one round or ``hop'' of message passing. Thus, a transformer with $M$ layers can only reliably propagate information $M$ positions; if the required dependency chain ($n$) exceeds $M$, exact computation is impossible within a single forward pass, regardless of width. This bottleneck is not mitigated by increasing hidden size. A wider model can increase per-message bandwidth but cannot reduce the minimum number of sequential rounds required for long-range dependency propagation~\cite{sanford2024transformers, peng2024on}.

\subsection{Long-Range Comparison Constraints}
Communication complexity theory establishes a range of lower bounds for tasks that require significant information relay across input positions. More generally, classic problems like substring equality, membership in complex formal languages, or enforcing cross-sequence consistency all become exponentially harder as the required number of rounds exceeds model depth. With limited layers, evidence from distant tokens is compressed into lossy summary representations, which leads to brittle, approximate answers and catastrophic errors at decision boundaries~\cite{chen2024multilayer}.

\paragraph{Example: Equality Testing Across Distant Substrings.} 
An archetypal communication bottleneck problem is checking whether two distant substrings in a long sequence are equal. Suppose we construct an input $S = u~\Vert~\text{filler}~\Vert~v$, where $u$ and $v$ are identical binary strings of length $n/2$, separated by substantial ``filler'' content. The task is to determine whether $u = v$~\cite{zhao2025llmsqlsolverllmsdeterminesql, wang2025stringllm}. In communication complexity, deterministic protocols for checking equality require transmitting $\Omega(n)$ bits between the parties holding $u$ and $v$. For transformers, self-attention should allow all-to-all interaction. In practice, computation is bottlenecked by the hidden dimension (bandwidth per token) and the number of available layers (communication rounds). For very long sequences, the information about $u$ must travel through hundreds or thousands of intervening filler tokens to reach $v$. This is analogous to a channel with severe bandwidth and round constraints. As separation grows, information about $u$ becomes diluted or corrupted. The model needs lossy summarization or hash-like approximations. While randomized protocols (e.g., hashing) can reduce communication, they introduce further error. Thus, transformers risk brittle failures and incorrect equality decisions as sequence length increases. Such effects are observed in long-context tasks, string matching, and symbolic reasoning benchmarks~\cite{bhattamishra2024separations}.

\subsection{Limitations of Sequential Augmentation}
CoT prompting and scratchpads extend the effective computation of fixed-depth transformers by spreading reasoning across additional generated tokens. Each token provides an intermediate step, partially compensating for the limited number of architectural rounds. Empirically, this improves performance on discrete reasoning tasks. Formally, such augmentation increases expressivity: scratchpads allow transformers to recognize richer languages than direct input–output mapping, but the benefit is bounded as the required compositional steps increase~\cite{merrill2023expressive}. 

\paragraph{Example: Multi-hop Relational Queries in Text.} 
Answering complex questions, like ``Is an entity mentioned in section~A referenced again under a different alias in section~C, and negated in section~E?'', requires integrating information across multiple distant spans. Each step is a ``hop'' along a chain of entities and mentions. Each self-attention layer enables one synchronous round of global communication, so resolving a $k$-hop relational query usually requires at least $k$ rounds. 
However, what if there are additional resources provided, like scratchpads, increased width, or specialized architectures?
Empirical studies still show that transformers often fail at multi-hop reasoning.
For example, \citet{peng2024on} demonstrates that increasing hidden size (bandwidth) does not compensate for insufficient depth (rounds) when logical structures are needed. 
Recent theoretical results~\cite{amiri2025lowerboundschainofthoughtreasoning} show that the length of the scratchpad must scale with task complexity.

\begin{tcolorbox}[colback=yellow!15!white, colframe=orange!80!black, title=Summary]
Communication complexity highlights that the limitations are fundamentally tied to the amount of information that can be exchanged across the model’s parallel structure. 
Above examples illustrate that transformer architectures are fundamentally limited not just by width but by lower bounds on sequential information relay. These constraints cannot be overcome by self-attention alone. Tasks requiring information to be propagated, compared, or composed across multiple long-distance dependencies most clearly expose these limitations. 
\end{tcolorbox}

\section{Open Questions and Future Directions}
\label{sec:discussion}


The convergence of circuit complexity, approximation theory, and communication complexity illuminates the boundaries of what current transformer architectures can achieve in discrete reasoning. The core findings reveal persistent barriers regarding exact symbolic computation, systematic generalization, and the reliable composition of step-wise logic. Here, we discuss the broader implications for the field, outline open questions, and suggest future directions toward more capable neuro-symbolic and state-aware reasoning systems.

\subsection{Implications of Theoretical Barriers}

To clarify how these frameworks reinforce each other, we summarize the implications and connections between each section:

\begin{enumerate}[itemsep=0pt, parsep=0pt]
\item \textbf{Depth–Communication Link}: Circuit complexity and communication complexity both show that transformer depth corresponds directly to the number of sequential computation or information relay steps: limited depth means both fewer computational layers (\emph{circuit}) and fewer rounds for propagating information (\emph{communication}) across input positions. Both views support that fixed-depth transformers have difficulty with long reasoning problems.

\item \textbf{Depth–Approximation Link}: The inability to dynamically add more layers to match problem complexity is related to approximation failures. As shown in approximation theory, functions with many discontinuities require either greater depth or exponentially sharper approximators. With fixed transformer depth, both frameworks conclude that errors will compound with each reasoning step.

\item \textbf{Approximation–Communication Link}: Approximation theory shows that transformers inherently smooth out their learned functions. Communication complexity demonstrates the difficulty in aggregating the distant pieces of evidence often needed for an exact answer. Thus, failures in exactness can be viewed as both a bottleneck in approximating discontinuities and in collecting all the information required for a reasoning problem.
\end{enumerate}

The above connections illustrate that the three frameworks are not independent. Transformer failures in discrete reasoning are best understood as the intersection and mutual reinforcement of all three frameworks.

\subsection{Open Research Questions}

While considerable progress has been made in understanding the limitations of transformers for discrete reasoning, a range of important questions remain open. Future research must address both architectural and training-level challenges to extend symbolic and systematic reasoning capabilities. Below, we highlight several central questions guiding exploration in this area:

\paragraph{How can we design architectures that enable symbolic reasoning beyond current transformer limitations?}
Many symbolic reasoning tasks fundamentally require precise, step-by-step computation and exact logical operations rather than the pattern-matching and smooth interpolations at which transformers excel. An open question is how to develop alternative model architectures, that can match or surpass the interpolation strengths of transformers while robustly supporting symbolic manipulation and exact algorithmic reasoning.


\paragraph{How can models develop an explicit awareness of decision boundaries and discontinuities?}
Current neural architectures treat most functions as continuous and tend to interpolate over decision thresholds, leading to errors in discrete or rule-based tasks. A major question remains: how can we train models to detect, represent, and reason about sharp boundaries and discontinuous changes, rather than smoothing over them~\cite{fengbird}?

\paragraph{Which training paradigms most effectively foster compositional generalization and multi-step reasoning?}
Even with expressive architectures, generalizing algorithmic behavior to new or longer compositions remains challenging. It is an active area of inquiry to determine what curriculum designs, intermediate supervision, data augmentations, or learning signals best encourage models to internalize the structure of multi-step, compositional rules and to robustly generalize this reasoning to novel or out-of-distribution inputs~\cite{lee2025self, cai2025extrapolation}.

\subsection{Recommendations for Future Architectures}

To address the above research questions, future work may look toward:
\begin{itemize}[itemsep=0pt, parsep=0pt]
    \item \textbf{Neuro-symbolic models:} Combining the pattern recognition strengths of neural networks with explicit, programmable symbolic components (e.g., logic circuits, formal parsers, or differentiable interpreters) may help preserve both algorithmic precision and flexibility~\cite{zhou2021hopper,ma2024exploring,calanzone2025logically}.
    \item \textbf{State-aware and memory-augmented architectures:} Mechanisms such as external memory~\cite{hatalis2023memory,hu2025memory}, structured state-passing~\cite{wu2024stateflow}, or state space models~\cite{gu2024mamba} could support unbounded sequential computation and long-range dependency management.
    \item \textbf{Boundary-aware learning and representations:} Developing models with inductive biases or objectives~\cite{ma2023graph, baheri2025hierarchical} that encourage awareness of sharp decision boundaries, symbolic transitions, or rule changes. The topic is related to manifold learning which has recently gained more attention in the LLM research community~\cite{xie2025mhc}.
\end{itemize}


Continued integration of machine learning, theoretical computer science, and symbolic AI research will be essential to overcome the intrinsic constraints of the current transformer paradigm.

\section{Conclusion}


The limitations of transformer-based models on discrete reasoning tasks arise from fundamental architectural properties, not merely from training or data deficiencies.
Theoretical insights from circuit complexity, approximation theory, and communication complexity together illuminate why transformers fail to reliably compose logical operations, propagate long-range dependencies, or approximate discontinuous decision boundaries. 

Our survey shows that these perspectives are not isolated and instead reinforce one another to reveal a consistent picture of discrete reasoning failures. While techniques such as chain-of-thought prompting or scratchpads offer partial relief, they do not fully overcome the inherent bounds imposed by the transformer’s design.
Addressing these barriers calls for principled innovations at the intersection of neural, symbolic, and sequential computation, moving beyond scaling alone toward more structurally expressive architectures.
By charting these theoretical boundaries and open questions, we hope to provide a clearer foundation for designing models that are not only more capable in discrete reasoning, but also more robust, interpretable, and generalizable across tasks that demand systematic, exact computation.

\section*{Limitations}
The survey's scope is limited to discussion on transformers in the perspective of three frameworks: circuit complexity, approximation theory, and communication complexity. It is very possible there may be alternative framings of transformer limitations that are not covered in this survey. Moreover, the survey's focus is on discrete reasoning tasks. We understand that there are many interesting tasks beyond discrete reasoning that also require more careful analysis. Finally, the survey may not cover all the variants of transformers in the existing literature. Due to space constraints, we highlight the most important and relevant work for the survey's theme.

\section*{Acknowledgments}
We immensely thank Dan Roth, Sujith Ravi, and Anna Rumshisky for their insightful comments and suggestions that helped improve the quality of this paper.
This work has been funded by Oracle but the views expressed in this paper are those of the authors. They do not reflect the policies or positions of Oracle or their affiliates. 
\bibliography{custom}

\end{document}